\pdfoutput=1

\documentclass[11pt]{article}

\usepackage[]{acl}

\usepackage{times}
\usepackage{latexsym}

\usepackage[T1]{fontenc}

\usepackage[utf8]{inputenc}

\usepackage{microtype}
\usepackage{array}
\usepackage{booktabs}
\usepackage{graphicx}
\usepackage{subfigure}
\usepackage{longtable}
\usepackage{multicol}
\usepackage{geometry}
\usepackage{tabularx}
\usepackage{xtab,afterpage}
\usepackage{caption}
\usepackage{times}
\usepackage{latexsym}
\usepackage{amsmath,amssymb,amsfonts}
\usepackage{multirow}
\usepackage{hyperref}
\usepackage{dblfloatfix}
\usepackage{verbatim}

\newcolumntype{L}[1]{>{\raggedright\let\newline\\\arraybackslash\hspace{0pt}}m{#1}}
\newcolumntype{C}[1]{>{\centering\let\newline\\\arraybackslash\hspace{0pt}}m{#1}}
\newcolumntype{R}[1]{>{\raggedleft\let\newline\\\arraybackslash\hspace{0pt}}m{#1}}

\usepackage{subfiles}

\usepackage{inconsolata}

\title{\textit{Explicit, Implicit, and Scattered:}\\ Revisiting Event Extraction to Capture Complex Arguments}

\author{Omar Sharif, Joseph Gatto, Madhusudan Basak,  Sarah M. Preum\\
Department of Computer Science, Dartmouth College \\
 \texttt{\{omar.sharif.gr, sarah.masud.preum\}@dartmouth.edu}}

\begin{document}
\maketitle

\begin{abstract}

Prior works formulate the extraction of event-specific arguments as a span extraction problem, where event arguments are \textbf{explicit} --- i.e. assumed to be contiguous spans of text in a document. In this study, we revisit this definition of Event Extraction (EE) by introducing two key argument types that cannot be modeled by existing EE frameworks. First, \textbf{implicit arguments} are event arguments which are \textit{not} explicitly mentioned in the text, but can be inferred through context. Second, \textbf{scattered arguments} are event arguments that are composed of information scattered throughout the text. These two argument types are crucial to elicit the full breadth of information required for proper event modeling. 

To support the extraction of explicit, implicit, and scattered arguments, we develop a novel dataset, \textbf{DiscourseEE},  which includes 7,464 argument annotations from online health discourse. Notably, 51.2\% of the arguments are implicit, and 17.4\% are scattered, making DiscourseEE a unique corpus for complex event extraction. Additionally, we formulate argument extraction as a \textit{text generation problem} to facilitate the extraction of complex argument types. We provide a comprehensive evaluation of state-of-the-art models and highlight critical open challenges in generative event extraction. Our data and codebase are available at \href{https://omar-sharif03.github.io/DiscourseEE/}{https://omar-sharif03.github.io/DiscourseEE}.
\end{abstract}





\section{Introduction}
Event Extraction (EE) is a challenging yet crucial NLP task required for event-centric information extraction. EE is the composition of two tasks: (i) Event Detection (ED), identifying \textit{if} an event occurs in a text and (ii) Event Argument Extraction (EAE), extracting event-specific details or event arguments according to a pre-defined event ontology. Existing works in EE have two key limitations. 

\begin{figure}[t!]
  \centering
  \includegraphics[width =\linewidth]{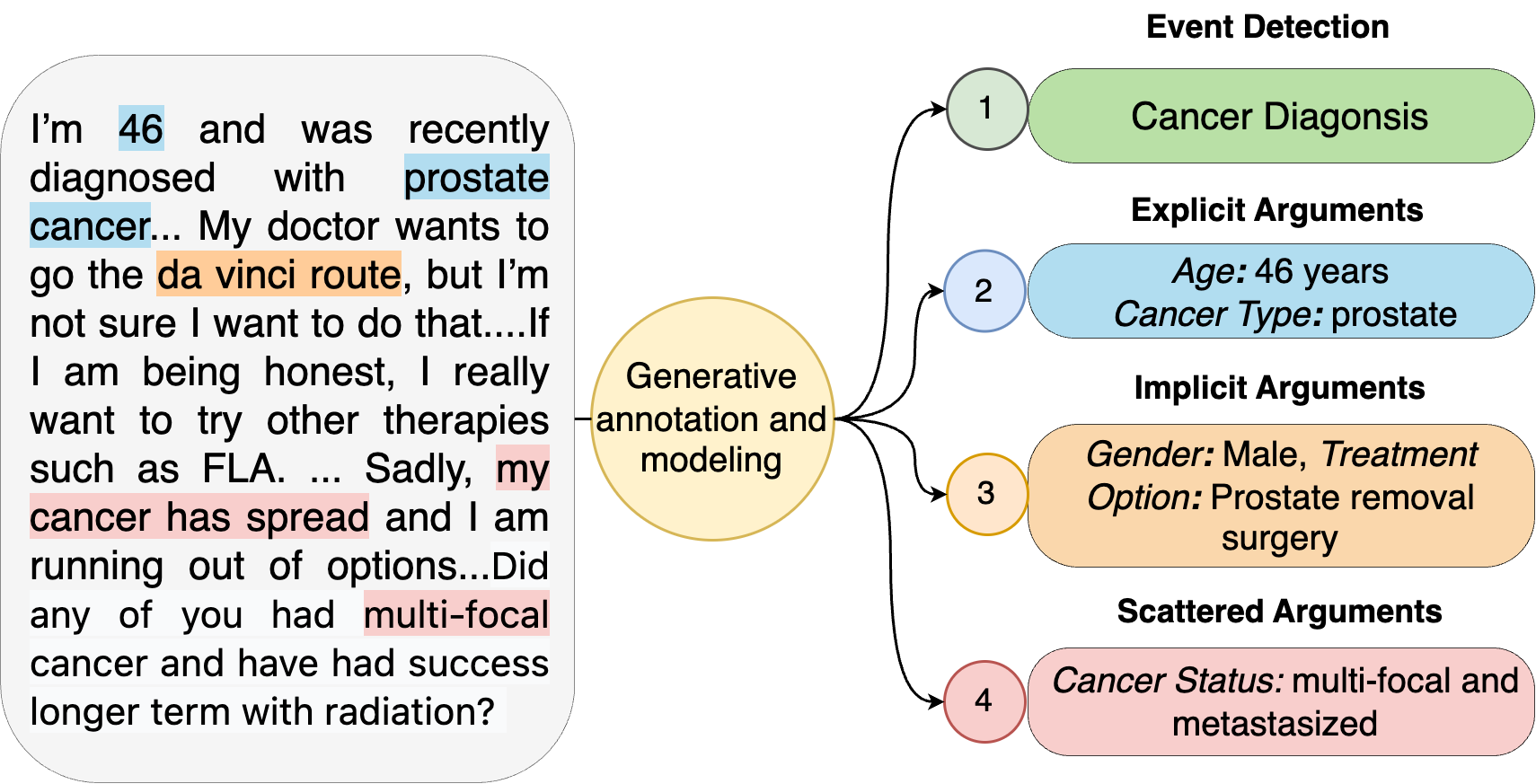}
 \caption{An example demonstrating complex event arguments that are prevalent in online discourse. This Reddit post is narrated by a newly diagnosed prostate cancer patient who is seeking treatment information from online peers on the r/ProstateCancer subreddit. In addition to explicit arguments, it contains
 implicit and scattered arguments that cannot be extracted using one contiguous span of text.}
 \label{intro-example-post}
\end{figure}

\textbf{First}, most prior works are focused on event extraction from formal texts (e.g., news or Wikipedia articles) \cite{doddington-etal-2004-automatic,tong-etal-2022-docee,du-cardie-2020-event}. This makes existing EE systems limited in their capacity to model other text sources such as social media or other colloquial text \cite{https://doi.org/10.48550/arxiv.2207.03997, lei-etal-2024-emona}. Insufficient data for EE on social media limits the ability of EE to facilitate downstream tasks such as mining online discourse \cite{jain-etal-2016-towards}, tracking dynamic events, knowledge base construction, rumor, and misinformation detection \cite{wu-etal-2022-cross}, moral value understanding \cite{zhang-etal-2024-moka}, or characterization of user behaviors \cite{rosa2020event}. 

\textbf{Second,} existing works in EAE extract an argument as \textit{a span in the input text} \cite{huang2024textee}. This is extremely limiting as many event arguments are implicit and only discernible from subtext. In Figure \ref{intro-example-post}, we illustrate the need for complex argument types through the lens of a Reddit post written by a newly diagnosed cancer patient. We find that only a small fraction of crucial event-specific details can be tied to contiguous spans of text in the post --- demanding a more complex EE solution. For example, the patient's consideration of prostate removal surgery is only discernible through a mention of the ``da vinci route", which is in reference to the Da Vinci Surgical Robot. Such implicit information is crucial to providing in-depth event details and will improve the accuracy of large-scale event-centric information aggregation efforts.

In this paper, we address these two limitations by reformulating EE annotation as text generation and introducing a novel dataset, \textbf{DiscourseEE} --- which contains EE annotation for health-related discourse on Reddit. By focusing on health discourse, we can provide a nuanced understanding of healthcare needs which has significant implications for public health research \cite{parekh2024event, guzman-nateras-etal-2022-event, romano2023theme, de2013predicting}. DiscourseEE introduces a novel EE annotation strategy that facilitates the extraction of the following three argument types: (i) \textit{\textbf{Explicit Arguments:}} event details that are found directly in the document. (ii) \textit{\textbf{Implicit Arguments:}} event details which are \textit{not} directly mentioned in the document but can be inferred using context. (iii) \textit{\textbf{Scattered Arguments:}}  event details which are the composition of multiple pieces of information scattered throughout the document.


While there is significant prior work on the extraction of explicit arguments, DiscourseEE is the first to introduce implicit and scattered argument extraction. When compared to prior work, our EE formulation adds significant depth to the amount of event information that can be extracted, making DiscourseEE well-suited to improve a range of downstream tasks such as question-answering \cite{jiang2024covid}, or rumor \cite{li-etal-2019-rumor}, conflicting information \cite{preum2017preclude2, preum2017preclude, Gatto_Basak_Preum_2023}, and misinformation \cite{wu-etal-2022-cross} detection from complex, online discourse. Additionally, by changing the EAE paradigm from span extraction to text generation, we better align EE with the abilities of Large Language Models (LLMs), which have been shown to have limited capacity on extractive EE tasks in prior work \cite{huang2024textee,gao2023exploring}. Our major contributions are as follows.



\begin{itemize}

    \item We introduce DiscourseEE, a dataset for characterizing event-level information in social media discourse. In addition to \textit{explicit} arguments, we introduce two prevalent yet overlooked argument types: \textit{implicit} and \textit{scattered}, broadening the scope of accessible knowledge in EE. DiscourseEE uses a novel event ontology, with 7,464 event-argument annotations leveraging relevant data from a health-related subreddit, i.e., a topic-specific community on Reddit. 51.2\% of arguments in DiscourseEE are implicit, and 17.4\% are scattered. To the best of our knowledge, this is the first large-scale, annotated social media dataset on event extraction with annotations for explicit, implicit, and scattered arguments.

    \item We reformulate EE annotation as a text generation problem to enable the extraction of non-explicit event information. We benchmark a diverse set of state-of-the-art event-extraction models on DiscourseEE, including both extractive models and several relevant LLMs. We identify limitations of existing models on DiscourseEE, motivating future works in EE. 
    
\end{itemize}

\begin{figure*}[h!]
  \centering
  \includegraphics[width =0.95\linewidth]{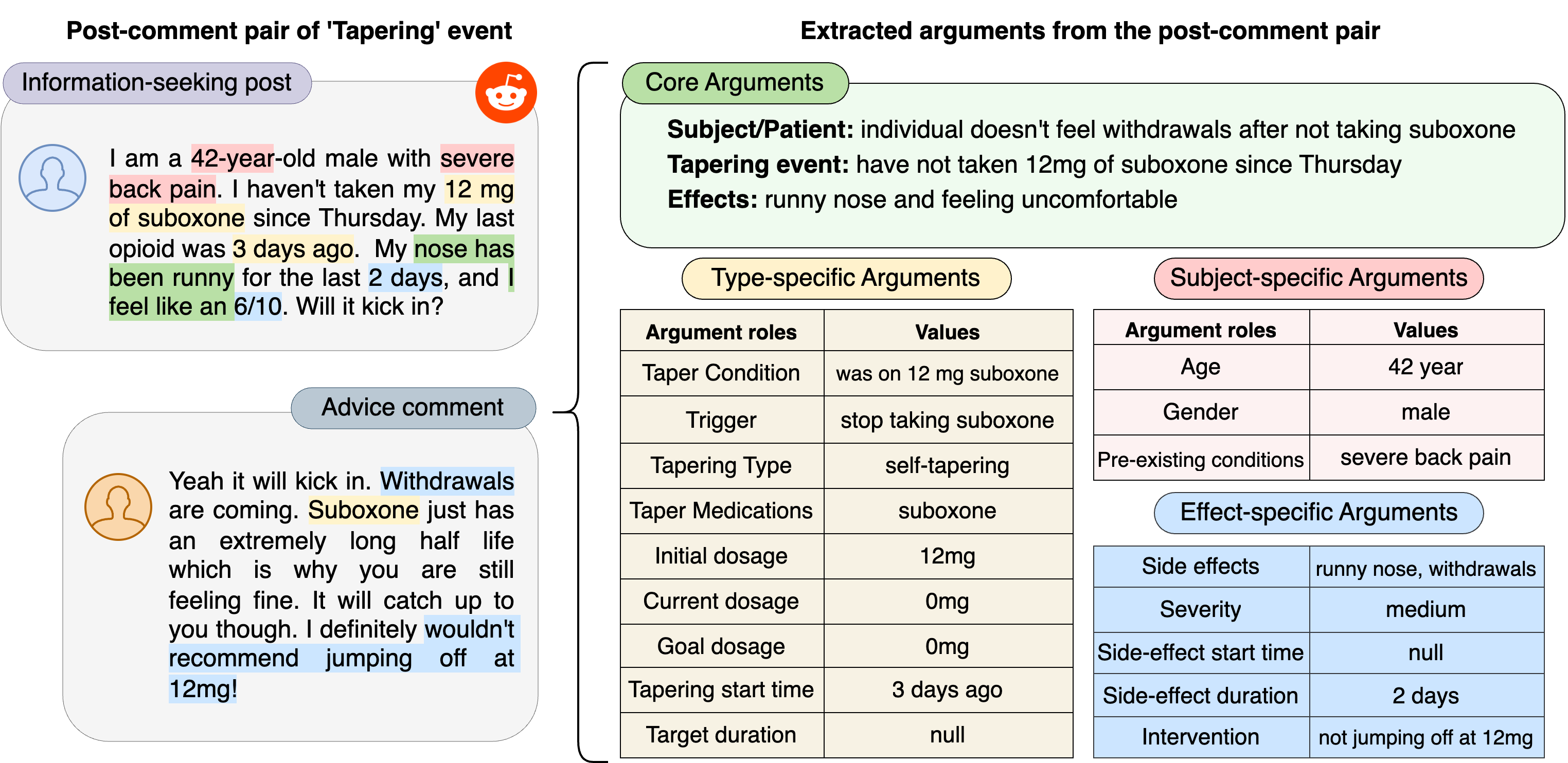}
  
 \caption{Example annotation in DiscourseEE. Core arguments capture the key aspects of the advice, while type-specific, subject-specific, and effect-specific arguments capture the fine-grained details. An argument can be explicit, implicit, or scattered throughout the document, e.g., as the individual is tapering suboxone, the \textit{goal dosage} is `0mg.' which is not directly mentioned in the text. We separately annotate the arguments from posts and comments. However, due to label sparsity, we merge them during model evaluation. The argument value is set to \textit{`null'} if absent, and multiple values for a role are comma-separated.}
 \label{sample-annotation-example}
\end{figure*}

\section{Event Extraction via Text Generation}
\label{section:EE-via-TG}

\textbf{Generative Event Argument Extraction (EAE):} Prior works on EE have exclusively focused on extracting arguments, which are  \textit{continuous spans} that can be found directly in the text \cite{du-cardie-2020-event,tong-etal-2022-docee}. In a real-world setting, this problem formulation limits one's ability to extract complex arguments such as those which are the composition of \textit{scattered information} throughout a text, or \textit{implicit information} with no direct mention in a text. In this study, we argue that implicit and scattered arguments are \textit{crucial} to understanding an event and that classic approaches to EE can not capture these arguments. To address this, we reformulate argument extraction as text generation rather than span extraction tasks and annotate arguments as natural texts.

\noindent
\textbf{Trigger-Free Event Detection (ED):}
Many prior works perform Event Detection (ED) by identifying event triggers ---  where the trigger is a word or phrase that best indicates the occurrence of an event \cite{du-cardie-2020-event,lu-etal-2023-event}. Recently, various studies used trigger-free ED, where texts are simply classified as containing an event without specific grounding to a trigger phrase \cite{tong-etal-2022-docee, liu-etal-2019-event}. We also adopt a trigger-free ED formulation as DiscourseEE events can be deeply implicit or the result of phrases scattered throughout a document, making it difficult to tie an event to a single trigger phrase.

\noindent
\textbf{Evaluating Generative Event Extraction (EE) Outputs:} A core challenge of implementing EE as a text generation problem is evaluating the quality of generated arguments. In prior EE formulations, all arguments correspond to start/end indices in a text --- thus, one can simply evaluate if the model has produced the `exact match' argument, i.e., identified the correct span \cite{huang2024textee}. Unfortunately, it is well established that `exact match' evaluation is not well-suited for models that generate human-like responses, such as LLMs \cite{wadhwa-etal-2023-revisiting}. Thus, if we attempt to translate the exact match evaluation to the generative setting, performance is severely underestimated as correct outputs can vary from ground truth. For example, if the ground truth argument for the role \textit{side effect} is ``\texttt{runny nose}", a generative model may correctly output one of \texttt{[runny nose, drippy nose, sniffly, nasal discharge]} yet only one answer would be accepted by exact match.



To solve this problem, we employ a relaxed match approach based on semantic similarity to achieve a more accurate evaluation. To account for variations in text, we consider the generated output and the ground-truth label for an argument to be similar if the semantic similarity score exceeds $0.75$. We compute BERT-based semantic similarity \cite{reimers2019sentencebert}. This threshold of 0.75 was determined through manual observation of the outputs. We acknowledge that changes in the threshold will affect the model's relaxed match score and suggest setting the threshold according to the downstream task. We also calculate the `exact-match' score for comparability with previous evaluations. We consider two sentences to be exactly matched when their semantic similarity is $1.0$. For evaluation, we calculate the F1 score based on both relaxed and exact matches and denote them as \textbf{RM\_F1} and \textbf{EM\_F1}, respectively.

\section{Event Ontology Design}
\label{event-type-arguments}

\begin{figure*}[h!]
  \centering
  \includegraphics[width =0.92\linewidth]{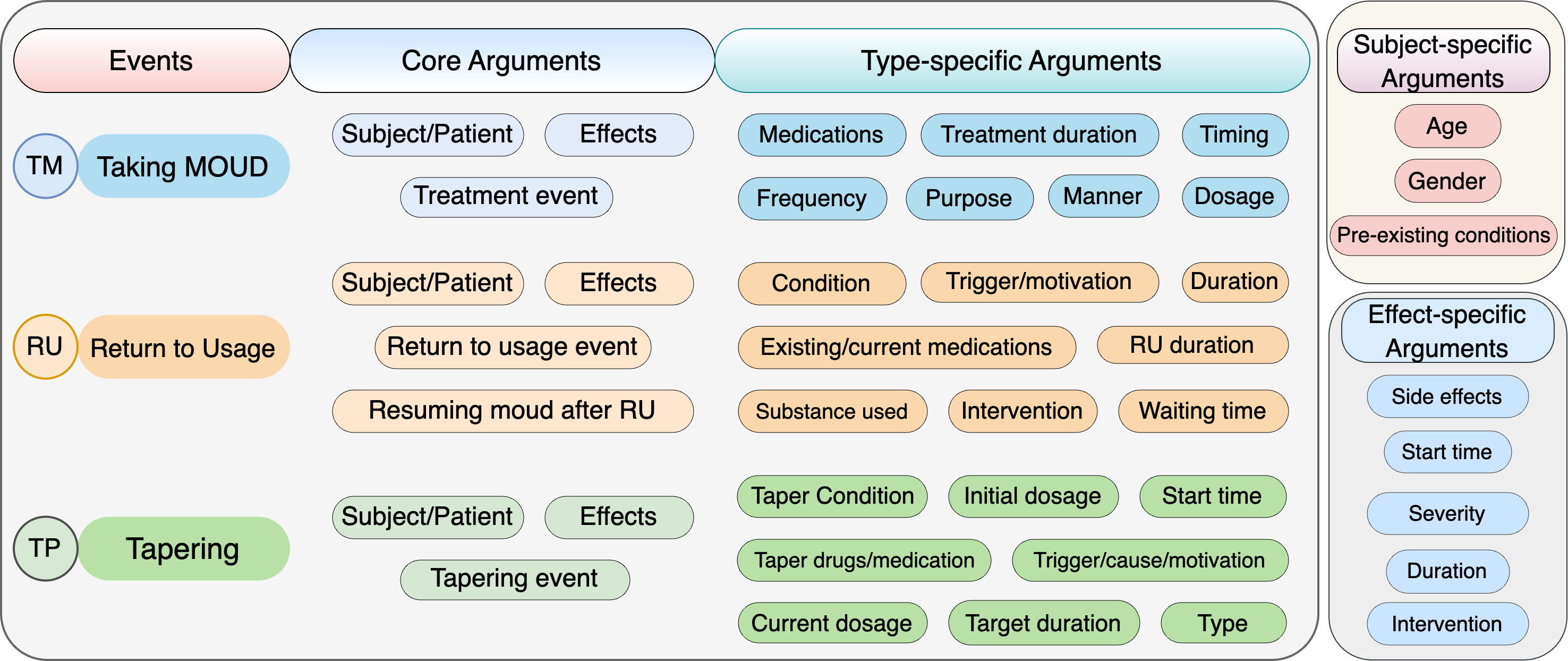}
 \caption{Event ontology of DiscourseEE dataset. Details of arguments provided in Table \ref{argument-details}.} 
 \label{event-arguments}
\end{figure*}

\textbf{Event Types:} 
Analyzing online discourse through an event-argument framework can inform data mining and knowledge discovery for several impactful domains including but not limited to healthcare, politics, public policy, finance, and law. As demonstrated in the motivating example in Figure \ref{intro-example-post}, millions of patients across the world seek informational and emotional support in online peer communities (e.g., Reddit, Facebook) on different conditions, e.g., mental health, pregnancy,  recovery from substance use disorder, cancer, and other chronic diseases. While our proposed method can be applied to any of these use cases, we found only one labeled, large-scale event dataset for online discourse / social media discussion, namely TREAT-ISE \cite{sharif2023characterizing}. This dataset covers health discourse regarding medications for opioid use disorder (MOUD), a critical yet stigmatized public health topic. OUD remains a leading cause of mortality in the US, incurring a massive economic toll, estimated at 1.02 trillion dollars annually \cite{FLORENCE2021108350}. There exists a lot of misperception and knowledge gaps regarding OUD treatment that impact treatment initiation and adherence. Thousands of affected individuals seek treatment information online among peers and make critical treatment decisions due to stigma, distrust on traditional healthcare, and lack of access to care. Event-driven analysis of such online discourse can inform MOUD-related public policy, patient communication and education. The dataset introduced by \citet{sharif2023characterizing} comprises five information-seeking events on medications for opioid use disorder (MOUD) from Reddit. These events include \textit{Accessing MOUD, Taking MOUD, Psychophysical effects, Relapse, Tapering}. 
We select three of these five event types as part of our event ontology: Taking MOUD (TM), Relapse (RL), and Tapering (TP). We exclude Accessing MOUD as they lack relevance to health advice. Additionally, we consider psychophysical effects as an event role instead of an event due to it's prevalence in all classes. In this work, we refer to `Relapse' as \textit{Return to Usage} (RU) in the rest of the paper as the former term can be stigmatizing\footnote{https://tinyurl.com/axtrbwrd}. Detailed descriptions of the relevant event types are provided in Appendix \ref{appnedix: event-ontology}.




\paragraph{Capture Complex Arguments} 

\noindent
We have a total of four types of arguments. 
\textbf{(1) Core Arguments:} longer texts containing the details of the advice event (e.g., subject receiving advice, advised treatment, outcomes or side effects of the treatment). The goal of annotating the core arguments is to get a high-level summary of the advice, which is difficult to infer from traditional annotations of short, discontinuous text spans. \textbf{(2) Type-specific Arguments:} words or phrases highlighting an advice event's specifics (e.g., advised treatment duration, medications). These arguments help to get a nuanced understanding of the specific event type. \textbf{(3) Subject-specific Arguments:} To understand advice, it is crucial to know to whom advice is given. These arguments are words or phrases providing details about the subject/patient (e.g., age, gender, prior medical history, or other social determinants of health). 
\textbf{(4) Effect-specific Arguments:} are words or phrases providing details about psychophysical effects (e.g., severity, duration) individuals experience or can experience that are related to an event, e.g., taking MOUD.

We defined \textbf{10 core, 23 type-specific, 3 subject-specific, and 5 effect-specific arguments} across three event types. Figure \ref{event-arguments} shows the event ontology we followed to develop DiscourseEE. To our knowledge, our dataset is the first discourse-level event extraction dataset enriched with fine-grained argument annotation capturing the real-world complexity of health advice on social media, including scattered spans and implicit arguments. Figure \ref{sample-annotation-example} shows a sample annotation of arguments for the `tapering' event. 
Short description of argument roles for each event type provided in Table \ref{argument-details}.

\section{DiscourseEE Dataset Curation}

\begin{table}[t!]
\centering
\small
\renewcommand*{\arraystretch}{1}
\begin{tabular}{L{3.8cm}|cc}
Step &\#Posts& \#Comments\\
\toprule        
Original dataset & 5,412 & 39,300\\
Filtering target events & 3,713 & 25,769\\
Filtering \textit{information-scarce} samples & 932 & 6,214\\
\bottomrule

\end{tabular}
\caption{Dataset summary at different stages of filtering}
\label{data-statistics-preprocessing}
\end{table}

\begin{table}[t!]
\small
\centering
\renewcommand*{\arraystretch}{1}

\begin{tabular}{L{3.6cm}|ccc}
& \textbf{TM}& \textbf{RU} & \textbf{TP} \\
\toprule

Avg. sample length (\#words) & 112.80 & 117.28 & 114.33\\
\#Arguments (without null arguments) & 1492 & 1213 & 1140\\
Avg. \#arguments per sample & 8.38 & 10.27 & 11.4\\
\midrule
\multicolumn{4}{c}{\# Explicit, implicit, scattered arguments}\\
\midrule
-- explicit & 462 & 449 & 295\\
-- implicit &756 & 546 & 668\\
-- scattered & 274 & 218 & 177\\

\bottomrule
\end{tabular}
\caption{DiscourseEE statistics across three event types. Here, TM, RU, and TP indicate `Taking MOUD', `Return to Usage', and `Tapering' events, respectively. Additional statistics of the dataset are shown in Table \ref{extra-dataset-statistics}.}
\label{dataset-statistics}
\end{table}

\subsection{Data Collection and Filtering}  



DiscourseEE expands the \textit{TREAT-ISE} dataset \cite{sharif2023characterizing,madhu_journal} which contains 5,412 information seeking Reddit posts on recovery treatment. We expand TREAT-ISE to include both \textit{information seeking} and \textit{information sharing} data by sourcing advice-centric comments on information seeking Reddit posts. We collect the 39,300 comments associated with TREAT-ISE dataset. We keep the posts/comments of our selected event types and discard the rest. We obtained 3,713 post threads comprising 25,769 comments and applied a two-step filtering process to discard information-scarce post-comment discussions. (1) \textbf{Sufficient Discourse Filtering:} We include only threads with at least 4 comments, ensuring a certain level of peer interaction. After excluding threads failing to meet this criterion, we have 2,432 posts with 23,101 comments. (2) \textbf{Discourse Length Filtering:} We removed threads where the initial posts (title and body) contained fewer than 10 words. We also removed comments with fewer than 10 words. We chose 10 as the filtering threshold here to ensure argument annotation quality as short samples do not have sufficient arguments. After this filtering process, we obtained 932 posts with 6,214 comments. These selected samples were then utilized for advice and event argument annotation. Table \ref{data-statistics-preprocessing} illustrates data statistics on different filtering stages.

\subsection{Identifying Information Sharing Content}


DiscourseEE aims to model event-centric information from social discourse, which contains information-seeking posts and information-sharing comments. Prior work has established how to source the former, but in this work, we introduce a method of automatically sourcing information sharing comments, i.e., containing advice. In other words, we aim to identify comments that provide \textit{answers} to information-seeking content. 



We follow the model in the loop \cite{chakrabarty-etal-2022-flute} annotation protocol for labeling post-comment pairs as advice followed by human-level verification. We apply different state-of-the-art open-source (Mistral) and close-sourced (GPT-4, GPT-3.5) LLMs to identify the advice in a comment. Note that since the context of the comment heavily depends on the post, we allow the model to view post content when determining if a comment contains advice. The GPT-4 model achieved the highest precision of 0.94, and we employ it to identify advice samples. Out of 6,214 post-comment pairs, the model categorized 2,934 as advice. This question-advice (post-comment) pair is utilized in the subsequent event argument annotation step. We manually validate the advice labeling accuracy of GPT-4 as discussed in Appendix \ref{appendix: advice-annotation}.

\subsection{Argument Annotation}
In DiscourseEE,  51.2\%  arguments are implicit, and 17.4\%  are scattered arguments. We focused on four types of arguments (\textit{core}, \textit{type-specific}, \textit{subject-specific}, \textit{effect-specific}) for each event. Figure \ref{sample-annotation-example} illustrates a sample annotation with associated argument values. Annotating such arguments takes more effort and time than classification tasks or span-based annotation. For the sake of feasibility, we randomly selected 396 post-comment pairs for argument annotation. Previously, researchers highlighted limitations of crowd-sourced annotation for complex tasks \cite{Zhang_Li_Zeng_Yan_Ge_2024}, including event detection 
 from online discourse \cite{parekh2024event,sharif2023characterizing}. Generative argument annotation in such data is even more challenging. It requires domain knowledge and interactive, progressive training sessions to ensure annotators understand the nuances. We decided to engage both domain experts and paid annotators recruited and trained locally at the authors' institution. 

Annotators have to write the values for each argument. For each sample, we annotated the comment and the corresponding post. The separated post and comment annotations are sparse. Therefore we merge the annotations and inspect both posts and comments as a pair. Each sample (i.e., post-comment pair) has 19 arguments on average. 

\textbf{Annotation process:} To complete the annotation, we formed a diverse group of 8 annotators: 4 graduate and 4 undergraduate students at the authors' institution. 
Each annotator underwent an extensive, four-week training period involving trial annotations to ensure proficiency in identifying event arguments and understanding the annotation guidelines. 
For each sample, at least two annotators wrote the \textit{core, type-specific, subject-specific,} and \textit{effect-specific} arguments. Note that all the arguments are annotated separately for each post-comment pair. The major challenges of such manual annotation include domain-specific terms and context, and identifying implicit and scattered arguments. These complexities are discussed in detail in Appendix \ref{appendix: annotation-complexity}. A third annotator reviewed each sample to correct potential errors and resolve any disagreement to ensure the reliability of the annotation process. 





\textbf{Inter-annotator Agreement:} 
Traditional Cohen's kappa is not suitable for our annotations due to the unknown number of disagreements. Following previous works \cite{sun-etal-2022-phee, thompson2018annotation}, we choose the F1 score to measure inter-annotator agreement (IAA). Specifically, we used relaxed match F1-score (RM\_F1) due to the generative nature of our annotation, which is discussed in Section \ref{section:EE-via-TG}. We have two sets of annotations for each sample. The F1 score is computed by selecting one annotation set as a `reference' to another. 
Through progressive training and interactive discussions, we achieve quality annotations (a 0.811 mean IAA score, which indicates substantial agreement). Finally, we have \textbf{DiscourseEE}, a novel discourse-level event extraction dataset comprising 7,464 event argument annotations.




\subsection{DiscourseEE Summary Statistics}

Table \ref{dataset-statistics}, shows the statistics of DiscourseEE. Our dataset differs from other general domain \cite{doddington-etal-2004-automatic} and clinical/pharmacovigilance \cite{ma-etal-2023-dice, sun-etal-2022-phee} EE datasets because of the higher average length ($\approx115$ words) and higher density of arguments per document ($\approx10$ words). Thus DiscourseEE is a reasonably sized EE dataset with fairly dense event arguments. We also annotated implicit and scattered arguments written as natural text, providing a novel resource for such argument (68.6\% of the total arguments) extraction. 
More dataset statistics are presented in Appendix \ref{appendix: data-statistics}.

\section{Benchmarking EE Models}

\begin{table*}[h!]
\small
\centering
\renewcommand*{\arraystretch}{0.95}

\begin{tabular}{l|ccc|ccc|ccc|c}

 & \multicolumn{3}{c}{\textbf{Taking MOUD}}& \multicolumn{3}{c}{\textbf{Return to Usage}} & \multicolumn{3}{c}{\textbf{Tapering}}& \\
\midrule
\textbf{Model}& C-A & TS-A & SE-A & C-A & TS-A & SE-A  & C-A & TS-A & SE-A & Mean (RM\_F1)\\
\midrule
Extractive-QA & 4.98 & 19.26 & 16.08 & 14.95 & 29.15 & 15.39 & 19.98 & 18.65 & 15.74 & 17.13\\ 
Generative-QA & &&& &&& &&& \\
-- FLAN-T5 (Base) &  34.99 & 37.22 & 11.40 & 25.37 & 20.31 & 11.37 & 38.78 & 40.94 & 17.53 & 26.44\\
-- FLAN-T5 (Large) & \underline{41.70} & 45.61 & 23.92 & 38.44 & 27.41 & 19.79 & \underline{44.04} & \underline{51.92} & 26.93 & 35.53\\
\midrule
\multicolumn{11}{c}{\textbf{LLMs with Zero-Shot Description-guided Prompt}}\\
\midrule
Gemma (7B) & 26.84 & 39.11 & 31.83 & 20.48 & 31.75 & 28.66 & 31.24 & 28.64 & 30.26 & 29.87 \\
 Mixtral (8x7B) & 34.19 & 30.70 & 33.94 & 33.77 & 33.73 & 31.80 & 40.55 & 41.51 & 39.47 & 35.52  \\
 Llama-3 (8B) & 32.88 & \underline{48.45} & 32.48 & 33.43 & 37.75 & 27.49 & 41.33 & 42.88 & 35.54 & 36.91  \\
Llama-3 (70B) & 41.35 & 39.39 & 25.28 & 35.20 & 38.80 & 30.78 & 41.54 &40.57 & 28.83 & 35.75 \\
GPT-4 & 37.88 & 46.34 & 30.50 & \underline{43.56} & \underline{41.94} & 39.68 & 42.90 & 38.43 & \underline{43.15} & 40.49 \\
\midrule
\multicolumn{11}{c}{\textbf{LLMs with Zero-Shot Question-guided Prompt}}\\
\midrule
Gemma (7B) & 25.52 & 46.46 & 29.00 & 15.66 & 33.51 & 24.94 & 27.63 & 36.39 & 32.38 & 30.16 \\
Mixtral (8x7B) & 36.89 & 27.88 & 31.31 & 32.53 & 33.06 & 20.44 & 33.67 & 32.95 & 42.69 & 32.38  \\
Llama-3 (8B) & 34.88 & 42.19 & 27.64 & 33.31 & 32.38 & 31.48 & 25.82 & 45.89 & 41.09 & 34.96 \\
Llama-3 (70B) & 37.28 & 42.31 & 25.02 & 35.94 & 41.42 & 33.49 & 36.13 & 40.28 & 34.90 & 36.31  \\
GPT-4 & 35.77 & 47.38 & \underline{38.83} & 39.75 & 40.41 & \underline{49.35} & 40.69 & 44.41 & 41.26 & \textbf{41.98}\\
\bottomrule 
\end{tabular}
\caption{Performance (avg. of 3 runs) of the models for event argument extraction across all argument types in relaxed match F1-score (RM\_F1). C-A, TS-A, and SE-A denote core, type-specific, and subject-effect arguments. The mean RM\_F1 is calculated by averaging the scores across all argument types for all three classes. The best score in each column is underlined. Model superiority is determined based on the mean RM\_F1 score. Models performance based on exact match F1-score (EM\_F1) presented in Table \ref{exact-match-argument-results}.}

\label{relaxed-match-argument-results}
\end{table*}

\begin{table}[t!]
\small
\centering
\renewcommand*{\arraystretch}{0.95}

\begin{tabular}{l|ccc}
& \textbf{P}& \textbf{R} & \textbf{F1} \\
\toprule
\multicolumn{4}{c}{\textbf{Transformer-based Models}}\\
\midrule
BERT &48.89  & 52.90  & 50.48 \\
MPNet & 47.91  & 65.88  & 54.95 \\
RoBERTa & 51.74  & 59.59  & 55.26 \\
\midrule
\multicolumn{4}{c}{\textbf{Instruction-tuned Models}}\\
\midrule
FLAN-T5 (base) & 54.12  & 53.48 & 51.26 \\
FLAN-T5 (large) & 57.61  & 54.78  & 55.63  \\
\midrule
\multicolumn{4}{c}{\textbf{LLMs with Zero-Shot Prompt}}\\
\midrule
Gemma-7B & 54.90  & 56.25  & 50.12\\
Mixtral-8x7B & 55.09 & 54.31 & 51.75 \\
Llama3-8B & 60.62 & 59.29  & 55.42 \\
Llama3-70B & 61.21 & 62.38  & 59.84 \\
GPT-4 & 62.36 & 64.62  & \textbf{61.40} \\

\bottomrule
\end{tabular}
\caption{Performance comparison (avg. of 3 runs) of the models for event detection (ED). P, R, and F1 indicate precision, recall, and macro-F1 scores, respectively.}
\label{ed-results}
\end{table}

In this section, we detail our experimental setup for benchmarking a wide range of models on ED and EAE for DiscourseEE. 

\subsection{Event Detection (ED) Models}
We formulate ED as a multilabel classification task, as each sample can provide information about multiple events. We employ three transformer-based models (BERT, RoBERTa, and MPNet), two instruction-finetuned models (FLAN-T5-base and FLAN-T5-large), and five large language models (Gemma-7B, Mixtral-8x7B, Llama3-8B, Llama3-70B, and GPT-4) for event detection. Our objective is to assess the feasibility of ED using large models and to examine the effects of scaling. Consequently, we experimented with models ranging from 7B to 70B parameters, including the closed-source GPT-4. These models demonstrated SOTA performance across various event extraction \cite{zhang2024ultra,wang-etal-2023-code4struct} and information extraction \cite{wadhwa-etal-2023-revisiting} tasks. Detailed descriptions of these models are provided in Appendix \ref{appendix: models}.

\subsection{Argument Extraction Models} We perform comprehensive experiments in three distinct settings: Extractive-QA, Generative-QA, and LLM-based generation with varying prompt types. In this set of experiments, we assume knowledge of the ground truth event type. We focus on a question-answering (QA) based approach as previous works achieved SOTA results with this model type \cite{du-cardie-2020-event,hsu-etal-2022-degree}. 

\textbf{Extractive-QA:} Following \cite{du-cardie-2020-event}, we implement a span extraction EAE baseline using question-answering. Specifically, the model is trained to map \texttt{(Question, Input Text)} $\rightarrow$ \texttt{(Argument)}, where each question is a function of the role we wish to extract\footnote{Questions used for this baseline can be found here: \href{https://tinyurl.com/44e8u5fx}{https://tinyurl.com/44e8u5fx}}. Note that since DiscourseEE is formulated as a generative task, we do not have span-level annotations. We thus use a BERT model pre-trained on general question-answering data \cite{rajpurkar-etal-2016-squad} to extract argument spans.


\textbf{Generative-QA:} We employ the instruction fine-tuning approach \cite{zhou2023lima} to develop the generative-QA models. The instruction set is created from the training data and the format is shown in Figure \ref{insturction-template}. To examine impact of model size on performance, we fine-tune two smaller models: FLAN-T5-base and FLAN-T5-large \cite{chung2022scaling}, as they contain less than 1 billion parameters.

\textbf{LLM-based Generation:}
\label{llm-based-generation}We conduct extensive experiments using both open-source and closed-source LLMs of various parameter sizes, including Gemma (7B), Mixtral (8x7B), Llama-3 (8B and 70B), and GPT-4. Models are evaluated in a zero-shot setting with two types of prompts: description-guided and question-guided. In the \textit{description-guided prompts}, role descriptions guide the models to extract arguments. In contrast, in the \textit{question-guided prompts}, questions are used to extract arguments similar to generative-QA and extractive-QA approaches. Since each event in DiscourseEE has an average of 19 arguments,   It will require a large number of inferences if we extract arguments for each role separately. On the other hand, extracting all arguments from a sample with only one inference (a) results in noisy outputs that are difficult to parse and (b) reduces accuracy.  So, to manage parsing complexity and inference costs, we employ a divide-and-conquer strategy. We note that \textit{subject-specific} arguments are rare in DiscourseEE. To manage experimental complexity we merge \textit{subject-specific} and \textit{effect-specific} arguments roles during evaluation. For the rest of the paper, we call it \textit{subject-effect} arguments. We extract \textit{core}, \textit{type-specific}, and \textit{subject-effect} arguments separately for each sample and then merge them.  The generic prompt template is illustrated in Figure \ref{fig-EAE-LLM-prompt}, with further prompt details in Table \ref{sample-prompts}. Details about each model, instruction prompt, fine-tuning, and hyperparameters are presented in Appendix \ref{appendix: models}, \ref{appendix:hyperparameter-details}.

\subsection{Experimental Setup}
\textbf{Data Splits:} DiscourseEE is partitioned into three mutually exclusive subsets: train (246 samples), validation (50 samples), and test set (100 samples). 
The same test set is used across all models for both tasks to ensure unbiased evaluation. All the training and fine-tuning experiments were done on the GPU-accelerated Google Colab platform. 

\noindent
\textbf{Prompt Setting:} Different LLMs require prompts and in-context examples optimized specifically for each model  \cite{ziems2024large}. In practice, users adopt a trial-and-error approach to find the optimal prompt for a model \cite{10.1145/3544548.3581388}. We evaluate a wide range of LLMs in a zero-shot setting to mitigate biases and reduce computational costs associated with finding the optimal prompts in few-shot setting. We use the same prompt across all models to (a) eliminate the confounding factor of prompt variation and (b) ensure a fair comparison of the models.

\noindent
\textbf{Performance Metrics:} We use macro F1-score to evaluate ED performance. For EAE, we employ the relaxed-match F1-score (RM\_F1) to identify the best models and also compute the exact match F1-score (EM\_F1). Scores are computed following prior work \cite{peng-etal-2023-devil}. The details of RM\_F1 and EM\_F1 are discussed in Section \ref{section:EE-via-TG}. Additionally, for EAE, we report the overall F1 score as well as the per-event type and per-argument type F1 in Table \ref{relaxed-match-argument-results}.



\section{Results and Discussion}
\textbf{Event Detection:} Table \ref{ed-results} illustrates the ED results, where GPT-4 achieved the highest F1 score of 61.40. Among the open-source LLMs, Llama-3 (70B) achieved the maximal score of 59.84. We notice a linear relation between model size and performance, with zero-shot performance improving as the model size increases. Interestingly, instruction-fine tuning enabled smaller FLAN-T5 models to achieve comparable performance (55.63). Conversely, the transformer models performed poorly, potentially indicating the complex nature of online discourse and this task. 

\begin{table}[h]
\small
\centering
\renewcommand*{\arraystretch}{1}

\begin{tabular}{l|ccc}

 & \multicolumn{3}{c}{\textbf{Relaxed Match (Recall)}}\\
\midrule
\textbf{Model}& Explicit & Implicit & Scattered \\
\midrule
Extractive-QA & 32.40 & 9.40 & 13.44  \\ 
Generative-QA &&& \\
-- FLAN-T5 (Base) & 35.49 & 23.72 & 33.33 \\
-- FLAN-T5 (Large) & 45.98& 34.15& 43.54 \\
\midrule
\multicolumn{4}{c}{\textbf{LLMs with Zero-Shot Description-guided Prompt}}\\
\midrule
Gemma (7B) & 37.96 & 24.13  & 28.31  \\
 Mixtral (8x7B) & 52.26 & 27.53  & 48.20 \\
 Llama-3 (8B) & 46.09 & 31.76  & 38.88  \\
Llama-3 (70B)  & 52.67  & 26.99  & 43.36   \\
GPT-4  &  53.39  & 33.46 & \textbf{54.12}  \\
\midrule
\multicolumn{4}{c}{\textbf{LLMs with Zero-Shot Question-guided Prompt}}\\
\midrule
Gemma (7B)  & 40.02  & 25.01  & 27.77  \\
Mixtral (8x7B) & 47.83  & 26.85  & 50.53 \\
Llama-3 (8B) & 41.35  & 30.67  & 36.91  \\
Llama-3 (70B)  & 53.49  & 28.35  & 41.21  \\
GPT-4  & \textbf{55.14}  & \textbf{36.53}  & 49.82  \\
\bottomrule 
\end{tabular}
\caption{Performance comparison of explicit, implicit, and scattered argument extraction. Model performance (avg. of 3 runs) is reported based on recall, showing how many explicit, implicit, and scattered arguments are extracted correctly.
}

\label{explicit-implicit-scattered-argument-results}
\end{table}


\noindent
\textbf{Event Argument Extraction:} Table \ref{relaxed-match-argument-results} shows model performance for argument extraction. GPT-4 with question-guided prompting attained the highest mean RM\_F1 score of 41.98. The instruction fine-tuned FLAN-T5 (large) model obtained a 35.53 score, outperforming several larger models. This indicates that instruction-tuned models achieve comparable performance when compute resources are limited. The extractive model performed poorly, achieving only a 17.13 RM\_F1 score, highlighting their limited scope in capturing implicit and scattered arguments. 

\noindent
\textbf{Explicit, implicit, and scattered arguments:} Table \ref{explicit-implicit-scattered-argument-results} compares the performance of various models in extracting explicit, implicit, and scattered arguments. All models achieved low scores in implicit argument identification, with the best-performing GPT-4 model reaching only 36.53. The extractive model performed poorly, identifying only 9.40\% of implicit and 13.44\% of scattered arguments. This weak performance is due to implicit arguments lacking direct mentions and scattered arguments consisting of discontinuous spans. 

\noindent
\textbf{Impact of exact-match evaluation:} We also evaluate models' performances using the `exact-match' (EM\_F1) metric. Due to space constraints, results are presented in the appendix (see Table \ref{exact-match-argument-results}, \ref{exact-match-explicit-implicit-scattered-argument-results}). 
These results show a significant performance drop in extracting core and subject-effect arguments, which are often implicit or scattered. To investigate further, we conducted a qualitative human evaluation on a subset of the best-performing GPT-4 outputs. The evaluation revealed that although the outputs are semantically similar, the exact-match evaluation frequently marked them as incorrect, underestimating the performance of the models. Addressing these evaluation issues is a crucial future direction for generative EE research.




\section{Related Work}

\textbf{Event extraction with generative models:} Prior studies have approached EE as a token-level classification or extractive task \cite{nguyen-etal-2016-joint-event, du-cardie-2020-event, wang-etal-2021-automated}. Recently, EE has been formulated as a text generation task using pre-trained language models, where the model is prompted to fill in natural language templates \cite{hsu-etal-2022-degree, lu-etal-2021-text2event}. With the advent of LLMs, generation-based EE gained more traction \cite{wang-etal-2023-code4struct, gao2023exploring}. However, most of these generative models are evaluated using span-based annotated datasets, which can underestimate their performance \cite{huang2024textee}. This is because the models' predictions may differ from the exact ground-truth spans yet still be accurate. This work addresses this gap by proposing a relaxed-match evaluation metric, and presenting a new dataset and benchmarks to facilitate generative event extraction. 

\noindent\textbf{Event extraction from social media:} Existing research on social media primarily addresses event detection and often overlooks argument extraction, a gap DiscourseEE addresses. For instance,  \citet{parekh2024event} detected epidemic-related events from tweets, while \citet{guzman-nateras-etal-2022-event} identified suicide-related events on Reddit. Arguments in social media data vary in span and are often ambiguous, implicit, and scattered. Our approach to implicit and scattered argument formulation differs from existing works. We define implicit arguments as those not directly mentioned in the document but can be inferred from the context and scattered arguments composed of information throughout the text. In contrast, existing works define implicit arguments as those that are 
\textit{explicitly mentioned but outside the fixed sentence window} of the event's trigger \cite{ebner-etal-2020-multi,zhang-etal-2020-two,liu-etal-2021-machine}. They overlook arguments that are entirely implicit or scattered. The event argument aggregation task done by \cite{kar-etal-2022-arggen} is aligned with our work, but they do not model implicit and scattered arguments. Moreover, they only focus on six discrete argument roles (Time, Place, Casualties, After Effects, Reason, and Participant) without grounding them with specific events. In contrast, our EE formulation adds significant depth to the amount of event information that can be extracted. We annotate 41 arguments from three events and characterize four types of complex arguments: core, type-specific, subject-specific, and effect-specific. Our formulation better captures the complexity and nuance of real-world EE applications.




\section{Conclusion}
This paper presents DiscourseEE, a discourse-level EE dataset with fine-grained annotations of complex event arguments and develops a novel pipeline for extracting these arguments. DiscourseEE provides a new resource for studying implicit and scattered arguments within complex social discourse, which has not been previously explored. The best performing GPT-4 model achieved only a 41.98\% overall F1 score, highlighting the challenges of extracting such intricate arguments. Specifically, the models accurately extracted only 36.53\% of implicit arguments and 54.12\% of scattered arguments, underscoring that effective argument extraction remains an open challenge. We believe DiscourseEE fills a critical gap in EE research and provides a valuable, timely dataset and benchmark for generative event extraction. 

\section{Limitations}

One limitation of our benchmarking effort is the reliance on relaxed matching (RM\_F1) to assess model performance. While we attempted to select an appropriate threshold by comparing model outputs with ground truth, some model outputs may still be inaccurate. Thus, we also report performance using the exact match approach (EM\_F1). However, EM\_F1 significantly underestimates model performance, highlighting the need for a more robust evaluation metric in future generative EE research.

DiscourseEE contains 396 annotated pairs with 7,464 arguments, which might seem small. However, this size is comparable to similar works with $\approx$ 8,000 arguments \cite{doddington-etal-2004-automatic, ma-etal-2023-dice}. Unlike previous works \cite{tong-etal-2022-docee}, our annotations require manually writing values for each argument instead of selecting spans, making scaling more difficult. We focused the use of our annotation budget on having high annotation quality, as this work lays the groundwork for future research in the domain. We engaged with domain experts, recruited students, and trained them for the annotation. Each sample is annotated by two annotators and reviewed by an expert. This rigorous process makes the scaling challenging and annotation expensive. In future work, we plan to extend the dataset size as more resources become available.

Finally, we evaluate all models in a zero-shot setting without tailoring prompts specifically for each one. While model-agnostic prompt optimization or alternative prompting techniques could improve performance, we did not pursue these experiments due to the high computational cost. Our focus is on benchmarking a broad range of models rather than optimizing a single model's output. Future research can explore few-shot learning, chain-of-thought prompting, and other techniques to increase model performance.

\section*{Ethical Considerations}
This research was approved by the author's institution's Institutional Review Board (IRB).

\paragraph{User Privacy:} All data samples were collected and annotated in accordance with the terms and conditions of their respective sources. No identifying personal information that could violate user privacy was collected or shared.

\paragraph{Biases:} Any biases in the dataset and model are unintentional. Data annotation was performed by experts and a diverse group of annotators following comprehensive guidelines, and all annotations were reviewed to mitigate potential biases. The developed dataset and model can only be used to detect events and identify arguments we discussed in the paper. The scope of using these resources for malicious reasons is minimal.

\paragraph{Intended Use:} We intend to make our dataset and models accessible to encourage further research on generative event extraction. 

\paragraph{Annotation:} Annotation was conducted by experts and trained student annotators. Key characteristics of our annotators include: (a) ages 22-30, (b) a mix of native and non-native English speakers, and (c) 1-5 years of research experience.  We provided detailed annotation guidelines,  including background knowledge of health advice, event extraction, and the type of information we wanted to extract to mitigate potential biases. All annotators were compensated as per the standard paying rate of the author's institution. 

\paragraph{Reproducibility} The model, parameter, and implementation details are presented in Appendix \ref{appendix: models}, \ref{appendix:hyperparameter-details}. Our code, evaluation and the dataset are available at \href{https://omar-sharif03.github.io/DiscourseEE/}{https://omar-sharif03.github.io/DiscourseEE}.

\bibliography{acl_latex}


\hfill

\newpage
\appendix
\section*{Appendix}

\section{Details of Event Ontology}
\label{appnedix: event-ontology}
We select three event types in DiscourseEE, and the definitions of selected event types are as follows. 

\paragraph{\textbf{Taking MOUD (TM):}}Events related to MOUD regimen details, e.g., advice about timing, dosage, frequency of taking a MOUD, suggestions about splitting and missing a dose. Analyzing advice from this event can surface potential misconceptions and concerns about MOUD administration that negatively impact treatment adherence.
\paragraph{\textbf{Return to Usage (RU):}} Events related to relapsing or using other substances during recovery. Such substance
    use can be attributed to recreational purposes or for self-medication (e.g., marijuana for sleep). 
    Advice in this event class can help unearth specific information individuals provide concerning recreational and medical usage of substances.
\paragraph{\textbf{Tapering (TP):}}Events related to reducing the dose or frequency of MOUD and eventually quitting MOUD. Although the current standard of care recommends consulting healthcare providers for tapering MOUD, individuals often resort to self-managed tapering strategies. Analyzing advice from this class can inform addiction researchers and clinicians about the context of self-managed tapering strategies (e.g., why and when people self-taper) and their effectiveness (what works for whom).

\section{Advice Annotation}
\label{appendix: advice-annotation}
 
\begin{figure*}[h!]
  \centering
  \includegraphics[width =0.9\linewidth]{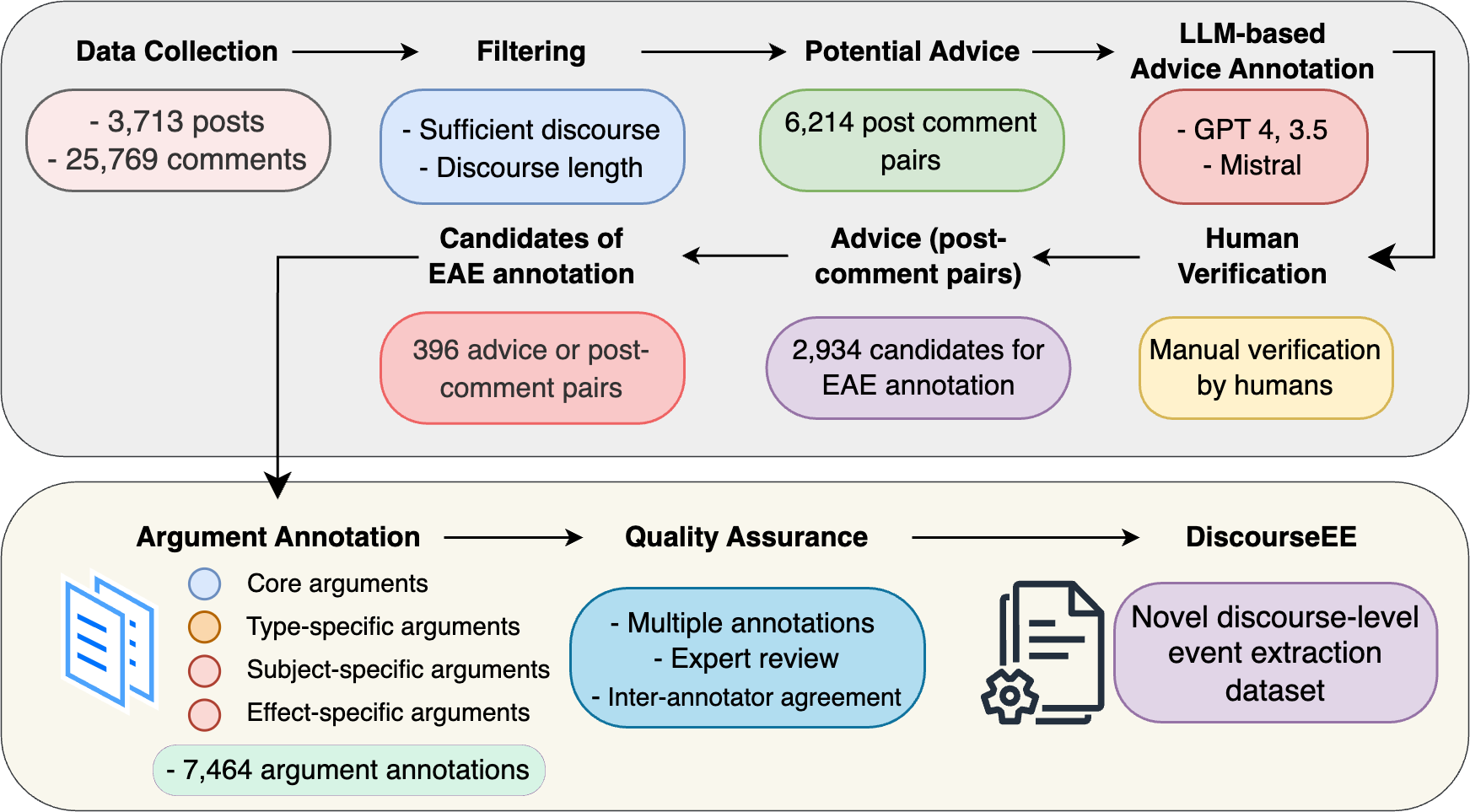}
 \caption{DiscourseEE development pipeline} 
 \label{dataset-development-pipeline}
\end{figure*}

 Figure \ref{dataset-development-pipeline} illustrates the DiscourseEE development pipeline. Recent works have indicated that LLMs, such as GPT and LLaMA, can be effective zero-shot data annotation tools \cite{he2023annollm}. LLMs have demonstrated the ability to reliably classify texts in various domains without supervision \cite{ziems2024large}. Therefore, we apply different state-of-the-art open-source (Mistral) and close-sourced (GPT-4, GPT-3.5) LLMs to identify potential advice. We developed an evaluation set of 100 samples (post-comment pairs) annotated by human experts as \textit{advice} and \textit{not-advice} to select the best model for our task. As our primary focus is on identifying advice, the \textbf{best model is selected based on the precision in the \textit{`advice'} class}. Table \ref{advice-results} shows the results of advice classification for various LLMs.

\begin{table}[h!]
\scriptsize
\centering
\renewcommand*{\arraystretch}{1}

\begin{tabular}{L{2.1cm}|ccc|ccc}

 & \multicolumn{3}{c}{\textbf{Advice}}& \multicolumn{3}{c}{\textbf{Not-Advice}}\\
\hline
\textbf{Model}& P & R&F1 & P & R & F1\\
\hline
GPT-4 (gpt-4- 1106-preview) & \textbf{0.94}  & 0.62 & 0.75 & 0.45     & 0.88  &    0.60\\

GPT-4 (gpt-4-0613) & 0.86 & 0.93 & 0.90 & 0.75  & 0.58   &   0.65\\

GPT-3.5 (gpt-3.5- turbo-1106) & 0.83 &  0.07  &    0.13 & 0.27  & 0.96  & 0.42\\


Mistral (7B-Instruct) & 0.85  & 0.64 &     0.73 & 0.40  & 0.69  &  0.51 \\
\hline            
\end{tabular}
\caption{Zero-shot advice classification of performance of different LLMs.}
\label{advice-results}
\end{table}

\begin{figure}[h!]
  \centering
  \includegraphics[width =0.9\linewidth]{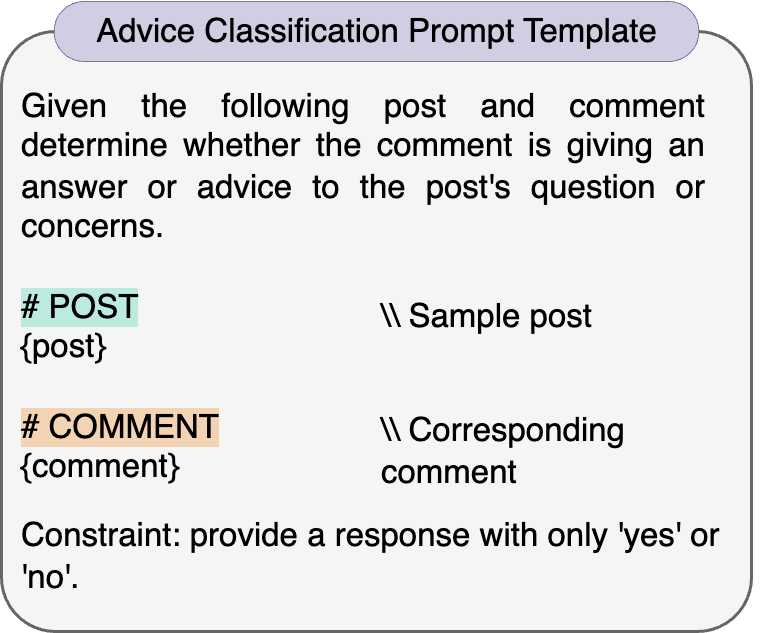}
 \caption{Advice classification prompt template.} 
 \label{prompt}
\end{figure}

We evaluate the zero-shot performance and do not tailor prompts specifically for each model. Instead, we write a simple prompt and use it for all the models.  Figure \ref{prompt} shows the structure of our prompt. The GPT-4 model achieved the highest precision of 0.94. While recognizing the potential for enhancing other models' performance through prompt optimization or the utilization of alternative prompting techniques (e.g., chain-of-thought, few-shot), we refrain from exploring these avenues due to the already high agreement observed between the human annotator and the GPT-4 model in the `advice' class. This aspect could be the focus of a separate study.

\paragraph{Final Annotation:} As the false positive is very low for \textit{advice} class, we employed the GPT-4 model to identify advice samples. Out of 6,214 post-comment pairs, GPT-4 categorized 2,934 as advice. This advice set is utilized in the subsequent event argument annotation step. 

\noindent
\paragraph{Human Verification:} To validate the labeling accuracy of GPT-4, we conducted a second-level verification. A human annotator manually reviewed 50 randomly selected samples labeled as `advice' by the model. The human annotator confirmed advice labels for 49 of the 50 samples, resulting in a 98\% precision for GPT-4 advice labeling.

\section{Annotation complexity}
\label{appendix: annotation-complexity}
To achieve quality annotations, we held biweekly meetings with the annotators to address challenges and complexities. Major challenges annotators face are, 

\begin{itemize}
    \item \textbf{Misleading context:} Commenters often share long stories of their treatment journey when giving health advice. Finding the correct arguments from such descriptions can be challenging. For instance, multiple mentions of \textit{`treatment dosage'} may make it difficult to differentiate between what is part of the current advice and what is simply mentioned about past conditions.
    
    \item  \textbf{Domain-specific terms and noisy short-hands:} Annotators struggle with understanding domain-specific terms like \textit{`Jumping off'} or \textit{`cold turkey'} which refer to quitting or the intention of not taking medications. Additionally, the presence of shorthand makes annotation challenging, such as \textit{`PWD'} for precipitated withdrawals and abbreviations like \textit{`percocet'}, \textit{`meth'}, \textit{`oxy'} representing different substances.   
    
    \item \textbf{Inconsistency in finding scattered arguments:} Arguments can be scattered throughout the document. For instance, for the core argument of \textit{`tapering event'}, components like \textit{`taper trigger,' `current dosage,' `taper start time'} may be dispersed all over. The chance of missing some parts of the arguments by annotators increased due to this dispersion. 
    
    \item \textbf{Implicit arguments:} Annotators face difficulties identifying implicit arguments as they require a deeper understanding of the context. Such as the severity of an individual's experience can be \textit{high, mild, low} or none. Understanding all psychophysical effects the individual is experiencing is crucial for selecting the severity when it is not explicitly mentioned.

\end{itemize}

\begin{figure*}[h!]
  \centering
  \subfigure{\includegraphics[width=0.45\linewidth]{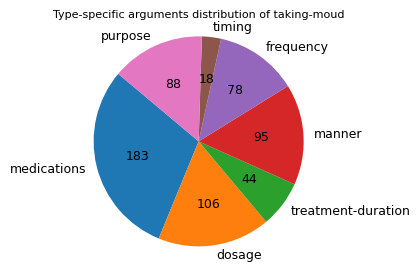}}
  \subfigure{\includegraphics[width=0.45\linewidth]{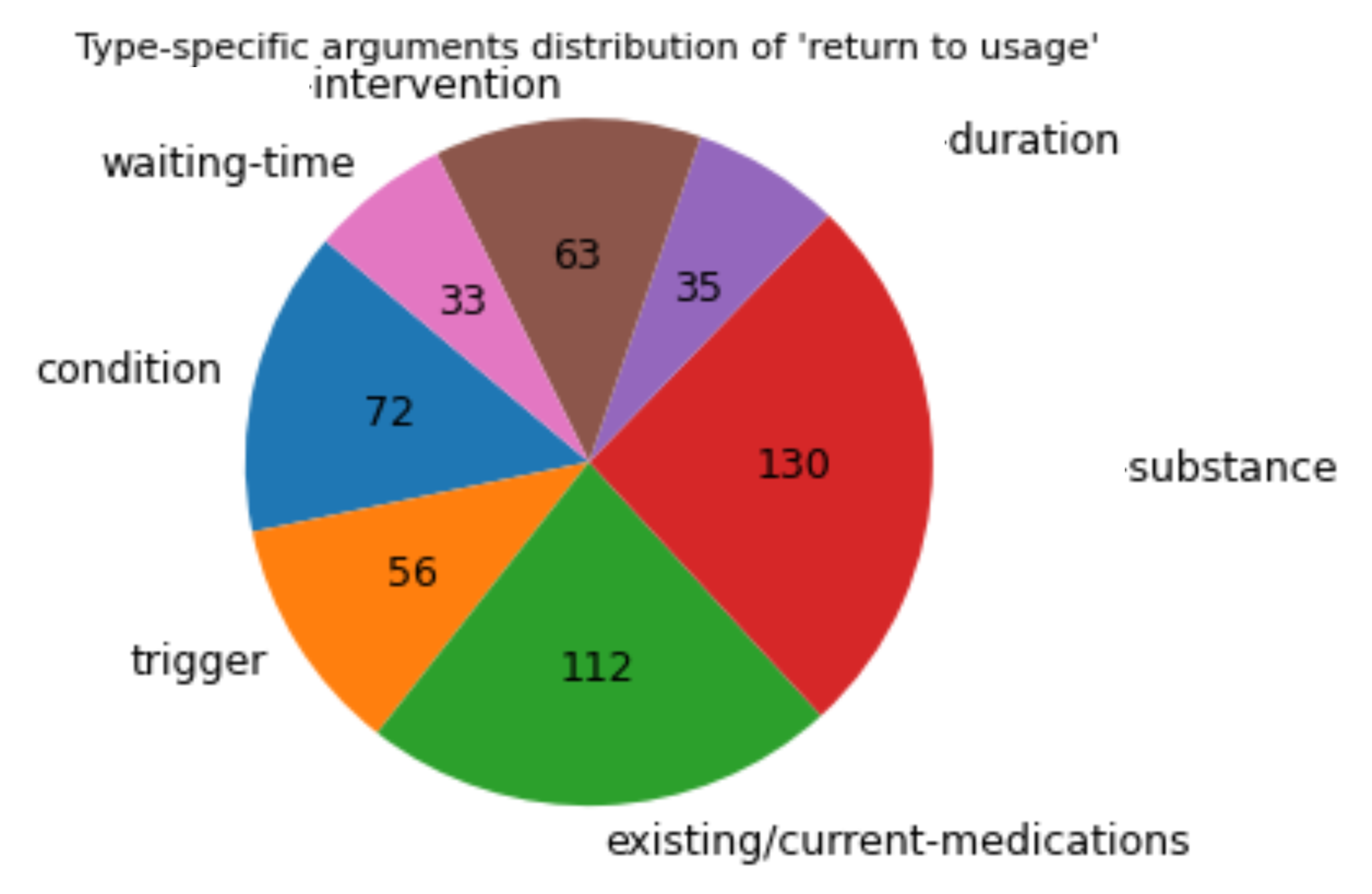}}
    \subfigure{\includegraphics[width=0.45\linewidth]{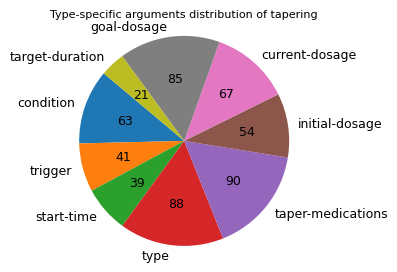}}
  \subfigure{\includegraphics[width=0.45\linewidth]{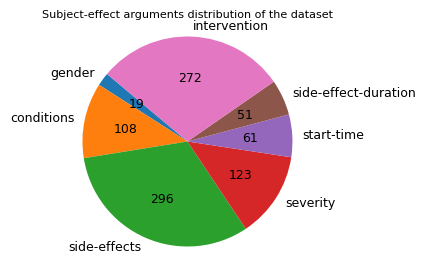}}

 \caption{Distribution of type-specific and subject-effect arguments in the dataset.} 
 \label{sub-argument-distribution}
\end{figure*}

We address these challenges to minimize annotation disagreements through interactive and iterative sessions, as well as multiple rounds of reviews.

\section{Dataset Statistics}
\label{appendix: data-statistics}

\begin{table}[h!]
\small
\centering
\renewcommand*{\arraystretch}{1}

\begin{tabular}{L{3.6cm}|ccc}
& \textbf{TM}& \textbf{RU} & \textbf{TP} \\
\toprule

Avg. \#Words per argument &&&\\
-- Core & 4.46 & 4.64 & 4.23\\
-- Type-specific & 1.70 & 1.95 & 1.93 \\
-- Subject-effect & 2.28 & 2.09 & 2.21\\
\midrule
\#Sample & 178 & 118 & 100\\
-- train & 117 & 67 & 62\\
-- dev & 20 & 16 & 14\\
-- test & 41 & 35 & 24\\
\midrule
\multicolumn{4}{c}{\textbf{Explicit, implicit, and scattered arguments}}\\
\midrule
\#Explicit arguments &&&\\
-- Core & 86 & 69 & 53\\
-- Type-specific & 252 & 266 & 151 \\
-- Subject-effect & 124 & 114 & 91\\
\midrule
\#Implicit arguments &&&\\
-- Core & 242 & 209 & 177\\
-- Type-specific & 321 & 190 & 346 \\
-- Subject-effect & 193 & 147 & 145\\
\midrule
\#Scattered arguments &&&\\
-- Core & 183 & 139 & 92\\
-- Type-specific & 39 & 45 & 51 \\
-- Subject-effect & 52 & 34 & 34\\

\bottomrule
\end{tabular}
\caption{Explicit, implicit, and scattered arguments distribution in DiscourseEE. }
\label{extra-dataset-statistics}
\end{table}
Table \ref{extra-dataset-statistics} shows the number of explicit, implicit, and scattered arguments across core, type-specific, subject-effect types. 68.6\% of the arguments are implicit or scattered, with only 31.4\% being explicit. The core arguments are predominantly implicit or scattered. Implicit arguments are more common in subject-effect categories, while type-specific arguments have a higher proportion of explicit ones.

Figure \ref{sub-argument-distribution} illustrates the distribution of type-specific and subject-effect argument annotations. For the `taking moud' event, mentions of \textit{medications, dosage}, and \textit{manner} are frequent. For `return to usage', people often mention their \textit{substance use, current-medications, condition} and \textit{trigger} of the return to usage. In `tapering', \textit{taper-medications, intial-dosage, current-dosage, } and \textit{type} of tapering are mostly mentioned. From subject-effect argument distribution, it is evident that people share more side-effect information (\textit{side-effects, intervention, severity}) than personal information (\textit{age, gender}). All these arguments provide crucial information for understanding or comparing health advice within social discourse.

\section{Models}
\label{appendix: models}

We perform comprehensive experiments encompassing various methodologies, including transformer-based, instruction-tuned, and large language models. The details of each model are described in the following.

\paragraph{Transformer Models:} We employ three transformer-based models to benchmark the event detection task. These include Bidirectional Encoder Representations from Transformers (BERT) \cite{devlin2019bert}, robust BERT architecture trained with more training data for longer period (RoBERTa) \cite{liu2019roberta}, and the model with permuted pre-training (MPNet) \cite{song2020mpnet}.

\paragraph{Instruction-tuned Models:} We choose FLAN-T5 \cite{chung2022scaling} as the backbone model for instruction fine-tuning. It follows the standard T5 \cite{raffel2023exploring} architecture and treats each task as a sequence-sequence problem. The model is fine-tuned on diverse task mixtures with instruction-following objectives and is available in a wide range of sizes, from small (80M parameters) to UL2 (20B parameters).  To explore the feasibility of our task with smaller language models and reduce computational costs, we experiment with the base (250M parameters) and large (780M parameters) models.

\paragraph{Large Language Models:} 
We used five LLMs for the experimentation.

\begin{itemize}
    \item \textbf{Gemma (7B)} \cite{gemmateam2024gemma} is an open-source language model trained on 6T tokens, following the architecture and training recipe of Gemini models \cite{geminiteam2024gemini}. It outperformed similarly sized open-source models on 11 out of 18 text-based tasks. Gemma comes in two sizes, 2B and 7B parameters. We utilize the 7B version for our experiments.
    \item \textbf{Mixtral (8x7B)} \cite{jiang2024mixtral} is a sparse mixture of expert language models with the similar architecture of Mistral 7B \cite{jiang2023mistral}. In mixtral, each layer comprises 8 feedforward blocks or experts, enabling each token to access 47B parameters while using only 13B active parameters. Due to this architectural change, mixtral outperforms models with higher parameters (e.g., Llama-2, GPT-3.5) across several benchmarks. 
    \item \textbf{Llama-3} is a state-of-the-art open-source language model pretrained and instruction-fined with 8B and 70B parameters. It builds upon the Llama-2 model \cite{touvron2023llama}, incorporating significant changes such as grouped query attention (GQA) and an expanded tokenizer vocabulary. The 8B and 70B models outperform others within a similar parameter range.

    \item \textbf{GPT-4} \cite{openai2024gpt4} is the best-performing multimodal model that achieved state-of-the-art performance on various professional and academic benchmarks. 
\end{itemize}

We employed the instruction-tuned version of all the models as it aligns better with our task. The HuggingFace inference strings for the open-source LLMs are Gemma (google/gemma-1.1-7b-it), Mixtral (mistralai/Mixtral-8x7B-Instruct-v0.1), and Llama-3 (8B) (meta-llama/Meta-Llama-3-8B-Instruct), Llama-3 (70B) (meta-llama/Meta-Llama-3-70B-Instruct). Additionally, we investigate the performance of the GPT-4 model via API (version gpt-4o-2024-05-13) calls.

\begin{table*}[h!]
\small
\centering
\renewcommand*{\arraystretch}{1}

\begin{tabular}{l|ccc|ccc|ccc|c}

 & \multicolumn{3}{c}{\textbf{Taking MOUD}}& \multicolumn{3}{c}{\textbf{Return to Usage}} & \multicolumn{3}{c}{\textbf{Tapering }}& \\
\midrule
\textbf{Model}& C-A & TS-A & SE-A & C-A & TS-A & SE-A  & C-A & TS-A & SE-A & Mean (EM\_F1)\\
\midrule
Extractive-QA & 1.25 & 7.83 & 1.50 & 3.36 & 14.85 & 6.47 & 5.11 & 4.20 & 4.00 & 5.40\\ 
Generative-QA & &&& &&& &&& \\
-- FLAN-T5 (Base) & 4.89 & 25.24 & 2.85 & 3.75 & 13.31 & 11.37 & 7.40 & 21.93 & 6.67 & 10.82 \\
-- FLAN-T5 (Large) & 5.10 & 32.33 & 7.87 & 7.09 & 18.57 & 14.66 & 10.06 & 33.58 & 10.81 & 15.56\\
\midrule
\multicolumn{11}{c}{\textbf{LLMs with Zero-Shot Description-guided Prompt}}\\
\midrule
Gemma (7B) & 2.35 & 20.88 & 14.21 & 1.11 & 16.33 & 12.65 & 8.45 & 17.76 & 12.57 & 11.81\\
 Mixtral (8x7B) & 7.25 & 18.34 & 11.41 & 3.80 & 13.12 & 12.17 & 8.79 & 19.01 & 11.68 & 11.73 \\
 Llama-3 (8B) & 2.34 & 28.88 & 9.79 & 4.42 & 19.05 & 13.86 & 8.19 & 25.88 & 15.24 & 14.18 \\
Llama-3 (70B) & 6.75 & 25.39 & 9.70 & 5.63 & 21.04 & 13.17 & 11.09 & 22.14 & 15.30 & 14.47 \\
GPT-4 & 2.36 & 27.30 & 14.79 & 7.39 & 23.04 & 21.04 & 8.42 & 20.35 & 16.77 & 15.72 \\
\midrule
\multicolumn{11}{c}{\textbf{LLMs with Zero-Shot Question-guided Prompt}}\\
\midrule
Gemma (7B) & 3.47 & 26.85 & 9.48 & 0.70 & 19.74 & 14.04 & 1.74 & 21.52 & 15.66 & 12.58 \\
Mixtral (8x7B) & 6.17 & 15.94 & 9.59 & 1.56 & 12.74 & 10.86 & 7.71 & 17.39 & 16.46 & 10.93 \\
Llama-3 (8B) & 1.85 & 21.72 & 6.99 & 4.92 & 18.88 & 11.77 & 5.81 & 25.75 & 14.31 & 12.45 \\
Llama-3 (70B) & 3.33 & 25.84 & 10.21 & 5.82 & 23.33 & 14.62 & 12.11 & 23.02 & 18.56 & 15.20 \\
GPT-4 & 4.95 & 29.06 & 19.35 & 6.22 & 21.19 & 26.10 & 8.72 & 21.06 & 14.86 & 16.84 \\
\bottomrule 
\end{tabular}
\caption{Performance (avg. of 3 runs) of the models for event argument extraction across all argument types in exact match F1-score (EM\_F1).
}

\label{exact-match-argument-results}
\end{table*}

\begin{table}[h!]
\small
\centering
\renewcommand*{\arraystretch}{1}

\begin{tabular}{l|ccc}

 & \multicolumn{3}{c}{\textbf{Exact Match (Recall)}}\\
\midrule
\textbf{Model}& Explicit & Implicit & Scattered \\
\midrule
Extractive-QA & 20.67 & 0.40 & 0.0  \\ 
Generative-QA &&& \\
-- FLAN-T5 (Base) & 26.54 & 10.02 & 1.61 \\
-- FLAN-T5 (Large) & 33.33 & 14.72 & 5.91 \\
\midrule
\multicolumn{4}{c}{\textbf{LLMs with Zero-Shot Description-guided Prompt}}\\
\midrule
Gemma (7B) & 26.74  & 7.43  & 0.89  \\
 Mixtral (8x7B) & 34.77  & 6.54  & 2.86 \\
 Llama-3 (8B) & 33.33  & 10.22  & 1.07  \\
Llama-3 (70B)  &  40.02  & 6.95  & 3.94  \\
GPT-4  & 37.96 & 10.77  & 3.22   \\
\midrule
\multicolumn{4}{c}{\textbf{LLMs with Zero-Shot Question-guided Prompt}}\\
\midrule
Gemma (7B)  & 30.86  & 8.17  & 1.25  \\
Mixtral (8x7B) & 31.06  & 7.56  & 2.68 \\
Llama-3 (8B) & 30.65  & 8.58  & 1.61 \\
Llama-3 (70B)  & 41.15 & 8.45  & 1.97 \\
GPT-4  & 39.81  & 11.86  & 3.04  \\
\bottomrule 
\end{tabular}
\caption{Performance (avg. of 3 runs) comparison of explicit, implicit, and scattered argument extraction in exact-match setting. Performance drop is significant in scattered arguments as these arguments are longer. Models generate different outputs with slight variation, which is not considered under exact match. 
}

\label{exact-match-explicit-implicit-scattered-argument-results}
\end{table}

\section{Implementation Details and Hyperparameters}
\label{appendix:hyperparameter-details}

Experimental details of our models are discussed in the following subsections. 
\subsection{Event Detection Models}
The transformer-based models (i.e., BERT, RoBERTa, MPNet) used for event detection are sourced from the HuggingFace library. All the models are fine-tuned on our training set with batch size $8$ and learning rate $2e^{-05}$. For each model, the version that achieves the best performance on the validation set is saved for final predictions on the test set.

Similar to transformer models, we also sourced the FLAN-T5 models from Hugginface. We use simple instructions for event detection. The input text template is \texttt{`Classify the following post-comment pair. [Post] [Comment]'} and the target text consists of corresponding event labels \texttt{`[Event labels]'}. Instructions sets are created from the training set. We fine-tuned base and large models for 3 epochs with a learning rate of $3e^{-4}$ and batch size 4. The max input length is set to 512 tokens, and the output length to 20 tokens.

We conduct open-source LLM experiments using LangChain\footnote{https://python.langchain.com/v0.1/docs/modules/ model\_io/prompts/} and HuggingFace. To generate event types with LLMs, we employ the template \texttt{`<instruction> <class details> <post> <comment>'}. Table \ref{sample-prompts} provides a sample prompt for event detection. This prompt template is utilized across all open and closed-source models.

\subsection{Argument Extraction Models}For argument extraction, we experimented with three suites of models. 
\noindent

\textit{\textbf{Extractive-QA:}} We implement the extractive-QA model using the HuggingFace pipeline with the BERT-base model (110M parameters) fine-tuned on the SQuAD dataset. The model is fine-tuned with a learning rate of $2e^{-05}$, a batch size of 8, for 3 epochs. We extract the argument for each role separately. During inference, we pass a question about the role along with the post-comment pair as the context. The input is formatted as \texttt{[CLS] Question [SEP] Post-comment pair [SEP]}. The output span is then decoded as the argument for the specific role.

\begin{figure}[h!]
  \centering
  \includegraphics[width =0.95\linewidth]{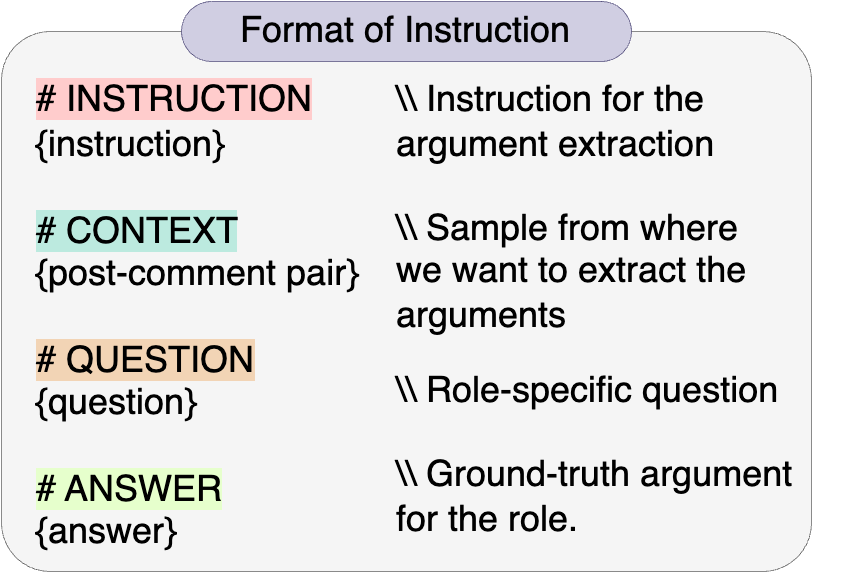}
 \caption{Instruction template for fine-tuning. See the sample instruction in table \ref{sample-prompts}.}
 \label{insturction-template}
\end{figure}

\begin{figure}[h!]
  \centering
  \includegraphics[width =\linewidth]{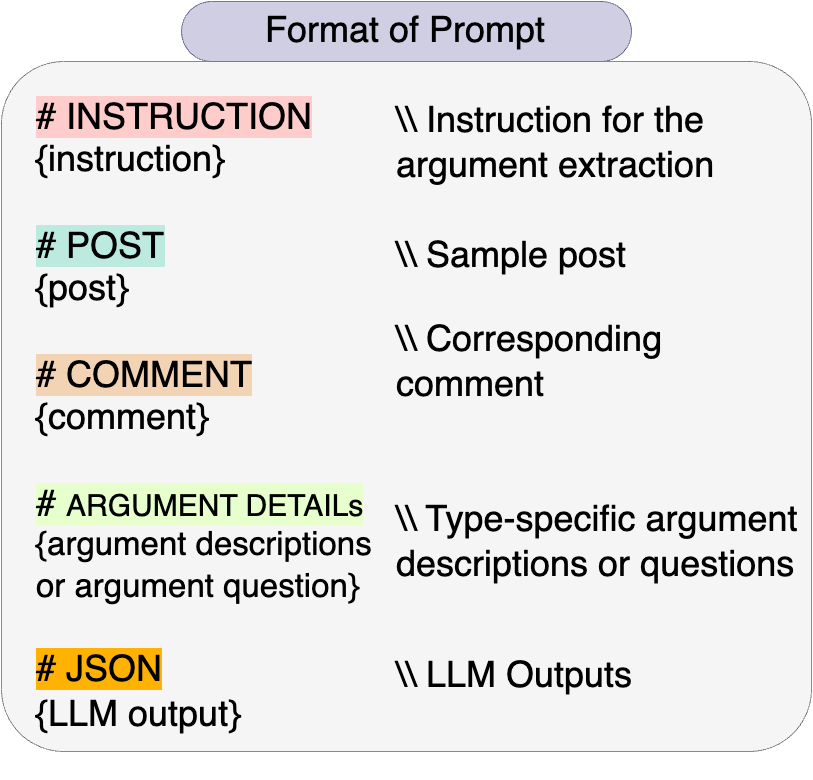}
 \caption{Generic argument extraction prompt template using LLMs.} 
 \label{fig-EAE-LLM-prompt}
\end{figure}
\textit{\textbf{Generative-QA:}} We use the HuggingFace \cite{wolf2020huggingfaces} Transformers library to develop the instruction fine-tuned generative-QA models. An instruction prompt is input to the model, which is fine-tuned to generate the argument for a role. Figure \ref{insturction-template} illustrates the format of the instruction prompt. Here, post-comment pair and answers all the values are filled from the training set. The model is fine-tuned for 2 epochs with a learning rate of $3e^{-04}$ and a batch size of 8. The input length is fixed at 512 tokens while the output length is 128 tokens. Similar to the extractive-QA approach, the argument for each role is generated separately. We fine-tune both the FLAN-T5-base (250M) and large (780M) models with the same hyperparameters on Google Colab A100 GPU.

\textit{\textbf{LLM-based Generation:}} Similar to event detection, we perform argument extraction experiments with LangChain and HuggingFace. We use instruction-tuned versions of the models as they align better with our task. To ensure the deterministic behavior of the model, we set the temperature value to 0.01 across all experiments with open-source models since it only allows non-zero values. The temperature is set to 0.0 for GPT-4. We experimented with all models using both description-guided and question-guided prompts. The prompt template for argument extraction is illustrated in figure \ref{fig-EAE-LLM-prompt}, and prompt examples are provided in table \ref{sample-prompts}.

\begin{table*}[h!]
\small
\centering
\renewcommand*{\arraystretch}{1}

\begin{tabular}{L{3cm}L{12cm}}
\toprule
Prompt template for event detection & \textit{\#Instruction:} Classify the following post into `taking-moud', `return to usage', `tapering' classes.
\textit{\#Class Descriptions:}\par
taking-moud: post related to medications for opioid use disorder (MOUD) regimen details.\par
 return to usage: post related to relapsing or using other substances during recovery. Such substance use can be attributed to recreational purposes or for self-medication (e.g., marijuana for sleep).\par
 tapering: post related to reducing the dose or frequency of MOUD and eventually quitting MOUD.\par
 \textit{\#Post}: <post> //post from the dataset.\par
  \textit{\#Comment}: <comment> //corresponding comment of the post.\par
 \textit{\#Output:} <LLM outupt> //LLM generated event types.\\
 
\midrule
Instruction template for fine-tuning FLAN-T5 models for argument extraction.  & \textit{\#Instruction:} Concisely extract the following argument from the post comment pair. Do not use more than 12 words to describe an argument. Return 'null' if any argument is not present.\par
\par
\textit{\#Context:}\par
Post: I am a 42-year-old male with severe back pain. I haven't taken my 12 mg of suboxone since Thursday. My last opioid was 3 days ago.  My nose has been runny for the last 2 days, and I feel like an 8/10. Will it kick in?
\par
Comment: Yeah it will kick in. Withdrawals are coming. Suboxone just has an extremely long half life which is why you are still feeling fine. It will catch up to you though. I definitely wouldn't recommend jumping off at 12mg!\par
\par
\textit{\#Question:}\par
What are the tapering steps (drugs, start dosage, duration, goal dosage)?\par
\textit{\#ANSWER:}\par
have not taken 12mg of suboxone since Thursday.\\
\midrule 
Description-guided argument extraction prompt template for LLMs. This is for `tapering' event `type-specific' argument extraction. &  \textit{\#Instruction:} Concisely extract the following argument from the post comment pair. Do not use more than 12 words to describe an argument. Return 'null' if any argument is not present. Return arguments in JSON format.\par

\textit{\#Post:} I am a 42-year-old male with severe back pain. I haven't taken my 12 mg of suboxone since Thursday.......
\par
\textit{\#Comment:} Yeah it will kick in. Withdrawals are coming. Suboxone just has an extremely long half life which is why you are still feeling fine. ....... \par
\textit{\#Arguments Descriptions:}\par
condition: Describe the state or situations of the subject before tapering,\par
trigger: Factors or events contribute to tapering,\par
start-time: Start-time of tapering,\par
type: Tapering type (self-tapering or prescribed tapering),\par
taper-medications: Drugs/medications used during tapering,\par
initial-dosage: Initial dosages of the drugs,\par
current-dosage: Current dosage of the drugs,\par
goal-dosage: Goal dosage the subject wants to achieve,\par
target-duration: Duration to go from the start to the intended dosage or quit.\par
\textit{\#JSON:}\\
\midrule

Question-guided argument extraction prompt template for LLMs. This sample is for `tapering' event `core' type argument extraction. &  \textit{\#INSTRUCTION:} Concisely extract the following argument from the post comment pair. Do not use more than 12 words to describe an argument. Return 'null' if any argument is not present. Return arguments in JSON format.\par

\textit{\#Post:} I am a 42-year-old male with severe back pain. I haven't taken my 12 mg of suboxone since Thursday.......
\par
\textit{\#Comment:} Yeah it will kick in. Withdrawals are coming. Suboxone just has an extremely long half life which is why you are still feeling fine. ....... \par
\textit{\#Arguments Questions:}\par
subject/patient: How can you describe the individual or patient involved?,\par
effects: What are the outcomes or side effects of the treatments?,\par
tapering-event: What are the tapering steps (drugs, start dosage, duration, goal dosage).\par

\textit{\#JSON:}\\
\bottomrule
\end{tabular}
\caption{Sample instruction and prompt used in the argument extraction experiments. To illustrate the difference in instruction, description-guided, and question-guided prompts, we used the same post-comment pair. For fine-tuned models, we extract arguments for each role separately. Thus, for each sample, we performed approximately 19 inferences (one for each role) based on the event type. To reduce inference costs for LLMs, we adopt a divide and conquer prompt approach (discussed in \ref{llm-based-generation}) .  We extract all arguments of a specific type (i.e., core, type-specific, subject-effect) together, reducing the number of inferences from 19 to 3. Finally, we merge the outputs to obtain the predictions.}
\label{sample-prompts}
\end{table*}

\begin{table*}[h!]
\small
\centering
\renewcommand*{\arraystretch}{1}
\small
\begin{tabular}{L{2.7cm} L{4cm}L{8.3cm}}

\toprule
\textbf{Type}&\textbf{Name} & \textbf{Description}\\
\midrule
\multicolumn{3}{c}{\textbf{Taking MOUD (TM)}}\\
\midrule 

\multirow{3}{*}{\texttt{Core Arguments}} & \texttt{Subject/Patient} & Describe the individual or patient involved. \\ 
 & \texttt{Treatment} & Describe the treatments prescribed or undergoing.\\
 & \texttt{Effects} & Describe the outcomes or side effects of the treatments.\\
\midrule

 {\texttt{Type-specific Arguments}}& \texttt{Medications} & Drugs/medications used in the treatment.  \\ 
 & \texttt{Dosage} & Current or previous dosage of the medications.  \\
 & \texttt{Treatment duration} & Duration of taking the medication. \\
  & \texttt{Manner} & Manner of taking medication orally/ sublingually/ as injections. \\
 & \texttt{Frequency} & Frequency of taking medication (per day, week, month) \\
 & \texttt{Timing} & Timing of taking medication (night, morning, etc.) \\
 & \texttt{Purpose} & Purpose of taking this medication. \\
\midrule
\multicolumn{3}{c}{\textbf{Return to Usage (RU)}}\\
\midrule
\multirow{4}{*}{\texttt{Core Arguments}} & \texttt{Subject/Patient} & Describe the individual or patient experiencing the return to usage.  \\ 
 & \texttt{Return to usage event} & Describe the occurrence of taking or using addictive substances. \\
 & \texttt{Resuming MOUD after RU} & Describe the events the subject is doing or intends to follow after the last return to usage dose. \\
 & \texttt{Effects} & Describe the outcomes or side effects of the return to usage. \\
 
 \midrule
{\texttt{Type-specific Arguments}} & \texttt{Condition} & Describe the substance use history/disorder from which the subject had previously recovered/ was in the process of recovery \\
 & \texttt{Trigger} & Factors or events contribute to return to usage.\\
 & \texttt{Existing/Current medications} & Medications used before the return to usage. \\
 & \texttt{Substance used in RU} & Substance was used in the return to usage. \\
 & \texttt{RU duration} & Duration of the return to usage. \\
 & \texttt{RU intervention} & Measures are taken to address or prevent the return to usage\\
 & \texttt{Waiting time} & Waiting time after the last dose of return to usage. \\
  \midrule

 \multicolumn{3}{c}{\textbf{Tapering (TP)}}\\
 \midrule

 \multirow{3}{*}{\texttt{Core Arguments}} & \texttt{Subject/Patient} & Describe the individual or patient involved.  \\ 
 & \texttt{Tapering Event} & Describe the tapering steps (drugs, start dosage, duration, goal dosage). \\
 & \texttt{Effects} & Describe the outcomes or side effects of the tapering. \\
 \midrule
 
{\texttt{Type-specific Arguments}} & \texttt{Taper condition} & Describe the state or situations of the subject before tapering. \\
 & \texttt{Trigger/motivation /cause/reason} & Factors or events contribute to tapering \\
 & \texttt{Taper Type} & Tapering type (self-tapering or prescribed tapering) \\
 & \texttt{Taper Drugs/ Medications} & Drugs/medications used during tapering \\
 & \texttt{Initial dosage} & Initial dosages of the drugs. \\
 & \texttt{Current dosage} & Current dosage of the drugs. \\
 & \texttt{Goal dosage} & Goal dosage the subject wants to achieve. \\
 & \texttt{Start time} & Start-time of tapering \\
 & \texttt{Target Duration} & Duration to go from the start to the intended dosage or quit.\\
\midrule 
 \multicolumn{3}{c}{\textbf{Taking MOUD / Return to Usage / Tapering}}\\
 \hline
\texttt{Subject-specific Arguments} & \texttt{Age} & Age of the subject/patient\\
& \texttt{Gender} & Gender of the subject/patient\\
& Pre-existing or comorbid conditions & Pre-existing or co-morbid conditions of the subject/patient\\
\midrule
\texttt{Effect-specific Arguments} & \texttt{Side Effects} & Side effects the subject is experiencing or expects to experience\\
& \texttt{Severity} & Severity of the side effects\\
& \texttt{Start time} & Start time of experiencing the side effects\\
& \texttt{Duration} & Duration of the side effects\\
&\texttt{Intervention} & Measures are taken to address or reduce side effects\\
\hline
\end{tabular}
 
\captionof{table}{\label{argument-details}Details of the argument roles for each event type in the DiscourseEE dataset. Subject-specific and effect-specific arguments are the same across all event types. }
\end{table*}

\end{document}